\documentclass{ceurart}

\usepackage{listings}
\begin{document}

\copyrightyear{2026}

\copyrightclause{Copyright for this paper by its authors. Use permitted under Creative Commons License Attribution 4.0 International (CC BY 4.0).}


\conference{EXPLIMED 2026 - Third Workshop on Explainable Artificial Intelligence for the medical domain - 15-17 August 2026, Bremen, Germany}

\title{From Continuous Predictors to Clinical Thresholds: Early Evidence on Performance Trade-offs of Guideline-Based Categorisation for Ischaemic Stroke Outcome Prediction}



\author[1]{Esra Zihni}[%
orcid=0000-0003-2288-2406,
email=esra.zihni@tudublin.ie,
url=https://github.com/esrazihni/,
]
\cormark[1]
\address[1]{Artificial Intelligence in Digital Health and Medicine, Technological University Dublin, Ireland}

\author[1]{Katryna Cisek}[%
orcid=0000-0002-2888-9943,
]

\author[2]{Hamzah Ziadeh}[%
orcid=0000-0001-9288-6520
]
\address[2]{Aalborg University, Aalborg, Denmark}

\author[2]{Hendrik Knoche}[%
orcid=0000-0003-3950-8453
]

\author[3,4]{Robert Mikulik}[%
orcid=0000-0002-7458-5166
]
\address[3]{International Clinical Research Center, St. Anne's University Hospital, Brno, Czechia}
\address[4]{Health Management Institute, Brno, Czechia}

\author[5]{John D. Kelleher}[%
orcid=0000-0001-6462-3248
]
\address[5]{Trinity College Dublin, Ireland}



\begin{abstract}
Machine learning models achieve strong predictive accuracy for 90-day outcome prediction in acute ischaemic stroke, yet clinical adoption is limited by the misalignment of model explanations with clinicians' reasoning. Motivated by a clinician user study calling for clinical guideline-aligned cut-offs, we ask whether continuous predictors can be replaced by clinically informed categorical encodings without sacrificing performance. On a multi-centre European registry stratified into three treatment cohorts, we compare standard and fully categorised gradient-boosted models, the latter using stroke guideline-aligned, treatment-specific thresholds. The fully categorised models are statistically indistinguishable from their continuous counterparts in two of the treatment cohorts, with a significant drop in predictive accuracy in one cohort. Global feature importance rankings remain consistent, suggesting that discretising continuous predictors into guideline‑based categories preserves the core hierarchy of prognostic factors across all treatment groups. Guideline-based categorisation is thus a viable design choice for stroke-outcome models.
\end{abstract}


\maketitle
\section{Introduction}
Machine-learning models for outcome prediction in acute ischaemic stroke have matured rapidly. Recent gradient-boosting and ensemble models, paired with SHapley Additive exPlanations (SHAP) \cite{lundberg_2017}, now routinely achieve AUROCs in the 0.81-0.91 range for 90-day functional outcomes on retrospective tabular data \cite{diprose_2024,abujaber_2025,chen_2025}. Predictive accuracy, however, is no longer the limiting step for clinical adoption. Recent studies of human-centred XAI evaluation in clinical decision support identify a different bottleneck: explanations that are technically faithful to the model but not aligned with clinicians' reasoning impose cognitive load, induce either mistrust or over-reliance, and fail to translate into actionable decisions \cite{sivaraman_2023,karagoz_2024}. This is the gap our work targets, in the specific setting of dichotomised 90-day mRS prediction for AIS.

Surveys of human-centred xAI evaluation suggest that meaningful explanations cannot be defined from inside the model and must be elicited from the intended end-users \cite{kim_2024}. In the clinical setting this elicitation typically combines qualitative methods, such as semi-structured interviews and think-aloud studies, with standardised quantitative instruments. To this end, we recently conducted a structured user study with 12 stroke clinicians, evaluating an earlier iteration of the standard models considered here. The study combined a quantitative arm (validated questionnaires covering causability, workload, and technology acceptance) with a qualitative arm (semi-structured interviews) in which participants interacted with individual SHAP explanations and a what-if interface. Eight of the twelve clinicians reported that continuous predictors (blood pressure, cholesterol, and blood glucose) added unnecessary detail and reduced their understanding of the model: small perturbations to raw values during 'what-if' analysis produced visible feature attribution changes that the clinicians considered clinically meaningless, since their own reasoning operates over guideline-aligned ranges rather than exact values. Six participants suggested training two models, one as presented to them in the study and one with clinically informed cut-off points to represent a prediction more aligned with their mental process. The implication of this feedback is that the limitation lies not in SHAP-based explanations themselves but in the units and reference frame in which feature attributions are presented to clinicians, an implication inferred in other recent studies \cite{panigutti_2022,hur_2025}.

Acting on the findings of this user study requires re-encoding continuous predictors as categories aligned with clinical guidelines. The biostatistics literature is notably cautious about such re-encoding since categorisation can result in loss of statistical power \cite{royston_2006}. Furthermore, most relevant work in clinical ML typically relies on data-driven cut-off points \cite{maslove_2013,chen_2019,tustumi_2022}, which makes the resulting models prone to systematic bias \cite{naggara_2011}. Clinically driven feature engineering, more broadly, has nonetheless been shown to reduce model complexity at little or no cost to accuracy \cite{roe_2020}, suggesting that categorisation driven by guideline-aligned thresholds may maintain predictive power while aligning explanations with clinical reasoning. Whether this is the case has not, to our knowledge, been examined.

In this work, we address this gap using 90-day outcome prediction in acute ischaemic stroke patients as a case study. We train gradient-boosted decision trees on a multi-centre stroke registry, stratified into three treatment cohorts. For each cohort we compare a standard model, in which continuous predictors are retained on their original scale, against a fully categorised model, in which every eligible continuous predictor is replaced by a categorical encoding whose thresholds are derived from stroke-specific clinical guidelines. Where guideline targets differ across treatment pathways, the same underlying predictor is encoded with treatment pathway-specific thresholds. To our knowledge, this is the first paired comparison of guideline-based categorisation of continuous predictors for 90-day mRS prediction in acute ischaemic stroke, motivated by a documented clinician user study rather than by statistical or computational considerations.

\section{Methods}

\subsection{Data and cohort}
We used a subset of the multinational RES-Q stroke registry data which included 81,735 records from Czechia, Bulgaria, Poland, Greece, Romania \cite{mikulik_registry_2017}.` Modelling was restricted to acute ischaemic stroke patients, and the prediction target was the 90-day modified Rankin Scale (mRS), which measures the degree of disability or dependence in daily activities. Cases were excluded if any of the following held: 90-day mRS missing; no imaging performed; discharge mRS = 6 (in-hospital death); or transfer to another hospital for treatment. Time-related predictors outside clinically relevant ranges (e.g., onset-to-door $\notin$ [0, 120] hours) and negative values for laboratory or numeric scoring predictors were removed, and IQR-based outlier filtering was applied to age (retained range 40-104 years). The final modelling cohort comprised 3,017 patients.

Since different treatment pathways represent distinct patient profiles, where different clinical and procedural factors influence outcomes, patients were stratified into three treatment cohorts, and a separate model was trained for each:
\begin{itemize}
\item \textbf{Cohort 1} — no recanalisation (n = 512)
\item \textbf{Cohort 2} — thrombolysis (n = 1,981);
\item \textbf{Cohort 3} — thrombectomy ± thrombolysis (n = 524);
\end{itemize}

\subsection{Predictors and preprocessing}
Predictors were chosen with clinical partners, prioritising predictors available before discharge so the model can support real-time decisions, and with an emphasis on modifiable factors. Predictors recorded at the time of discharge (e.g., discharge NIHSS, discharge medication) were intentionally excluded from this iteration. The candidate set covers baseline demographics and risk factors, prior medication, baseline labs and vitals, imaging, treatment workflow times, and early post-treatment findings (full list in Appendix A).

For each treatment cohort, predictors with >10\% missing values or with a dominant category (most-frequent category > 100× the combined frequency of all others) were removed. This yielded 28 / 30 / 32 predictors for cohorts 1 / 2 / 3, respectively. Missing values were imputed within each train/test split using median/mode imputation computed from the training set only, to avoid leakage.

\subsection{Guideline-based categorisation of continuous predictors}
The contribution of this work is replacing the standard continuous predictors with clinically grounded categorical encodings chosen to align with actionable decision thresholds clinicians use at the bedside. Categorised predictors span baseline severity and demographics (age, NIHSS), metabolic markers (blood pressure, blood glucose, cholesterol), and workflow times (onset-to-door, door-to-imaging, door-to-needle, door-to-groin, door-to-reperfusion). Cut-offs for metabolic markers were taken from AHA/ASA guidelines \cite{prabhakaran_2026}, whereas relevant stroke studies were used to decide on the age and NIHSS cut-off points \cite{adams_1999,mortensen_2020}. Workflow-time encodings reflect graded performance targets; for door-to-needle, for instance, we used $\leq$15 / 15-30 / 30-45 / >45 min, capturing quality-improvement benchmarks motivated by evidence that each additional 15-min door-to-needle delay is associated with worse one-year outcomes. Where guidelines differ by treatment pathway (notably blood pressure and onset-to-door), the same predictor is encoded with treatment pathway-specific thresholds across the three cohorts. The full set of thresholds is summarised in Table~\ref{tab:thresholds}.

\begin{table}[h]
\centering
\caption{Guideline-aligned thresholds used to categorise continuous predictors. Treatment pathway-specific encodings are indicated in the \emph{Cohort(s)} column. BP: blood pressure.}
\label{tab:thresholds}
\begin{tabular}{@{}llp{6.2cm}@{}}
\toprule
\textbf{Predictor} & \textbf{Cohort(s)} & \textbf{Categories} \\
\midrule
Age (years)                & 1, 2, 3 & $<$65 / 65--$<$80 / $\geq$80 \\
NIHSS score                & 1, 2, 3 & $<$6 / 6--$<$16 / $\geq$16 \\
\midrule
\multirow{2}{*}{Systolic BP (mmHg)}    & 1    & $\leq$110 / 110--$<$220 / $\geq$220 \\
                                       & 2, 3 & $\leq$110 / 110--$<$180 / $\geq$180 \\
\midrule
\multirow{2}{*}{Diastolic BP (mmHg)}   & 1    & $<$120 / $\geq$120 \\
                                       & 2, 3 & $<$105 / $\geq$105 \\
\midrule
Blood glucose (mg/dL)      & 1, 2, 3 & $<$60 / 60--140 / $>$140 \\
Cholesterol (mg/dL)              & 1, 2, 3 & $<$70 / 70--100 / $>$100 \\
\midrule
\multirow{3}{*}{Onset-to-door (h)}     & 1 & $\leq$24 / $>$24 \\
                                       & 2 & $\leq$4.5 / 4.5--24 / $>$24 \\
                                       & 3 & $\leq$6 / 6--24 / $>$24 \\
\midrule
Door-to-imaging (min)      & 1, 2, 3 & $\leq$20 / $>$20 \\
Door-to-needle (min)       & 2       & $\leq$15 / 15--30 / 30--45 / $>$45 \\
Door-to-groin (min)        & 3       & $\leq$60 / 60--90 / $>$90 \\

Door-to-reperfusion (min)  & 3 & $\leq$90 / 90--120 / $>$120 \\
\bottomrule
\end{tabular}
\end{table}

We refer to the models with continuous predictors retained on their original scale as \textbf{standard models} and to the new variants (all eligible continuous predictors re-coded as in Table~\ref{tab:thresholds}) as \textbf{fully categorised models}. All other predictors and modelling choices are held fixed across the two variants, isolating the effect of guideline-based categorisation.

\subsection{Multicollinearity}
Gradient-boosted trees are robust to multicollinearity in terms of predictive performance, since each split selects the most informative feature from the candidate predictors, but high inter-feature correlation can still distort SHAP-based importance attribution. Because feature importance is central to this work, we ran a multicollinearity check on the full predictor set of every cohort × variant combination (six predictor sets in total). We computed the adjusted squared GVIF, which extends the standard VIF to categorical predictors with multiple levels and uses the same interpretive threshold (>10 indicates problematic collinearity) \cite{fox_monette_1992}. Across all three cohorts and both the standard and fully categorised variants, every adjusted squared GVIF was < 5, so no problematic multicollinearity is present in any of the predictor sets, and SHAP attributions can be interpreted without correction.

\subsection{Prediction target}
The prediction target is the dichotomised 90-day mRS (favourable, mRS $\leq$ 2, vs. unfavourable, mRS > 2). The class imbalance varies across the three treatment cohorts:
\begin{itemize}
\item \textbf{Cohort 1} — no recanalisation: favourable / unfavourable $\approx$ 332 / 180 ($\approx$ 2 : 1)
\item \textbf{Cohort 2} — thrombolysis: favourable / unfavourable $\approx$  1453 / 528 ($\approx$ 3 : 1)
\item \textbf{Cohort 3} — thrombectomy ± thrombolysis: favourable / unfavourable $\approx$  312 / 212 ($\approx$ 1.5 : 1)
\end{itemize}

No resampling, class weighting, or threshold tuning was applied; both standard and fully categorised models are trained on the same class distributions.

\subsection{Models and hyperparameter tuning}
All six models (3 cohorts × 2 variants) were developed using CatBoost gradient-boosted decision trees, chosen for its native handling of categorical predictors and built-in overfitting control \cite{dorogush_catboost_2018}. Hyperparameters were tuned by 5-fold cross-validated grid search on each training set: tree depth $\in$ {4, 6, 8}, learning rate $\in$ {0.01, 0.03, 0.1}, L2 regularisation rate $\in$ {3, 10, 100}. The number of trees was set automatically by CatBoost's overfitting detector. 

\subsection{Performance estimation and final model training}
\begin{figure}[h]
  \centering
  \includegraphics[width=0.7\linewidth]{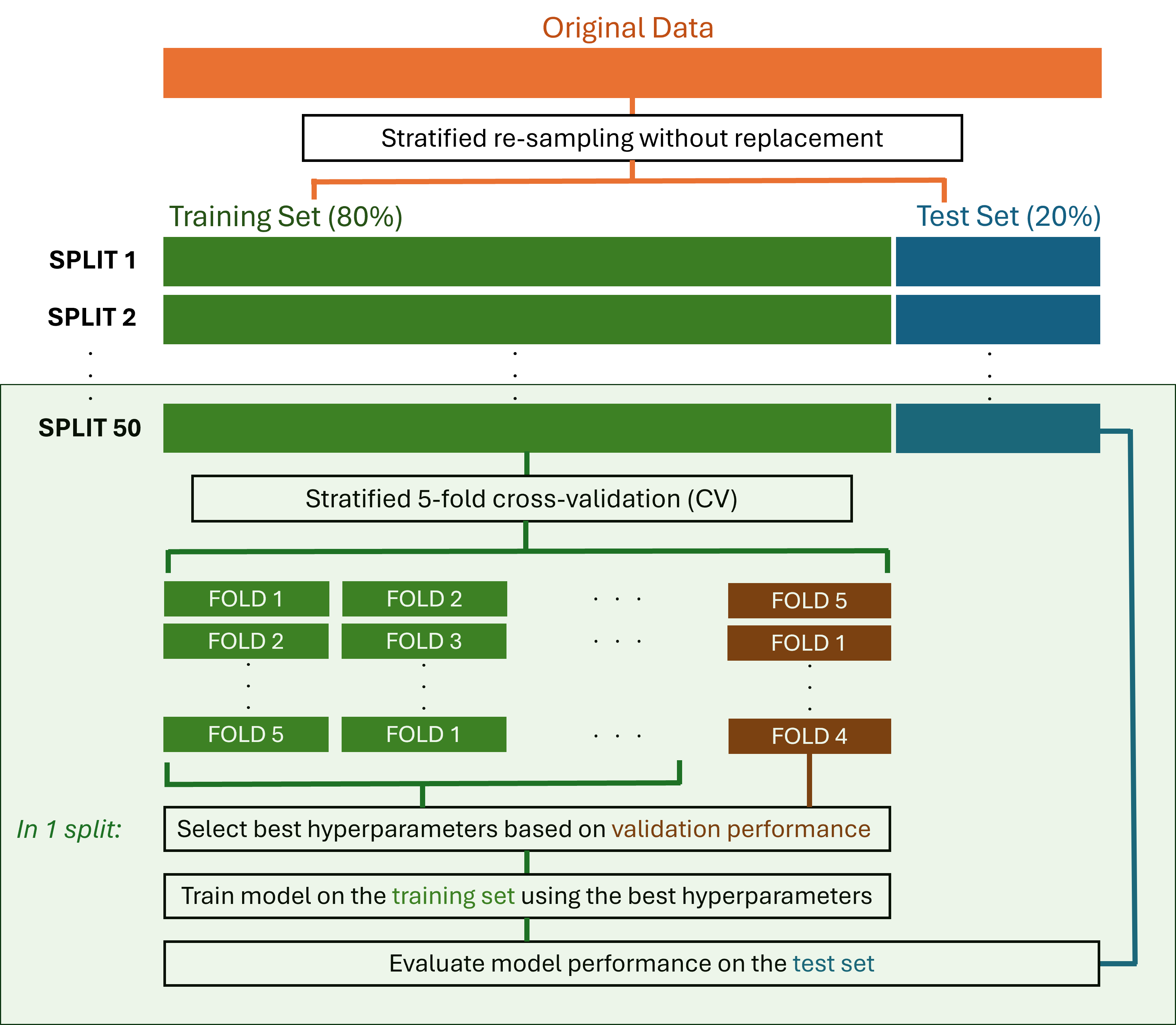}
  \caption{Performance-estimation workflow. The data is randomly partitioned into training and test sets 50 times, where 5-fold CV is used to optimise hyperparameters on each training set. The resulting optimal models are then evaluated on each test set.}
  \label{fig:experiment_design}
\end{figure}

Generalisation performance was estimated via Monte-Carlo cross-validation with 50 random 80/20 train/test splits. The hyperparameter tuning processed described in the previous section was repeated for the training partition of each split. The workflow is summarised in Figure~\ref{fig:experiment_design}.

Final models were trained on the complete available dataset for each cohort, using hyperparameters set to the mean or mode of the values obtained across the 50 tuned training splits (Table \ref{tab:hyperparameters}). SHAP values were subsequently computed for each of these final models.

\begin{table}[h]
\centering
\caption{Hyperparameters of the final standard and fully categorised models across the three treatment cohorts.}
\label{tab:hyperparameters}
\begin{tabular}{@{}lllp{2cm}@{}}
\toprule
\textbf{Hyperparameter} & \multicolumn{3}{l}{\textbf{Value (Standard/Fully Categorised)}} \\
 & \textbf{Cohort 1} & \textbf{Cohort 2} & \textbf{Cohort 3}\\
\midrule
Number of trees & 186/171 & 194/127 & 214/158 \\
Tree depth & 6/4 & 4/4 & 4/4 \\
Learning rate & 0.03/0.03 & 0.03/0.1 & 0.03/0.03 \\
L2 regularisation rate & 3/3 & 10/100 & 3/3 \\
\bottomrule
\end{tabular}
\end{table}

\section{Results}

\subsection{Generalisation performance}
We report the mean test-set metric across the 50 splits, with 95\% confidence intervals, using AUROC, balanced accuracy, F1 score, sensitivity, and specificity (Table~\ref{tab:performance}). Differences between the standard and fully categorised models are tested with the Wilcoxon signed-rank test at $\alpha$ = 0.05. Across the three cohorts, the fully categorised models retain most of the discrimination of the standard models, with a significant difference observed only in the thrombolysis cohort.

\begin{table}[h]
\centering
\caption{Mean test-set performance (95\% CI) over 50 Monte-Carlo splits. Asterisks mark metrics where the paired difference between standard and fully categorised variants is significant (Wilcoxon signed-rank, $p<0.05$).}
\label{tab:performance}
\setlength{\tabcolsep}{4pt}
\small
\begin{tabular}{@{}llccccc@{}}
\toprule
\textbf{Cohort} & \textbf{Variant} & \textbf{AUROC} & \textbf{Bal.\ Acc.} & \textbf{F1} & \textbf{Sensitivity} & \textbf{Specificity} \\
\midrule
\multirow{1}{*}{No recanalisation}
  & Standard    & 0.820          & 0.714          & 0.725          & 0.514          & 0.913 \\
  &             & (0.809--0.831) & (0.703--0.724) & (0.714--0.736) & (0.491--0.537) & (0.903--0.923) \\
  & Fully & 0.802          & 0.693          & 0.702          & 0.491          & 0.895 \\
  & Categorised & (0.788--0.815) & (0.679--0.707) & (0.687--0.718) & (0.466--0.516) & (0.882--0.908) \\
\midrule
\multirow{1}{*}{Thrombolysis}
  & Standard    & 0.807*          & 0.662*          & 0.682*          & 0.386*          & 0.938 \\
  &         & (0.802--0.813) & (0.656--0.668) & (0.675--0.688) & (0.375--0.397) & (0.934--0.942) \\
  & Fully & 0.785*         & 0.634*         & 0.650*         & 0.324*         & 0.945 \\
  &   Categorised          & (0.779--0.790) & (0.629--0.640) & (0.643--0.657) & (0.311--0.336) & (0.941--0.949) \\
\midrule
\multirow{1}{*}{Thrombectomy $\pm$}
  & Standard    & 0.812          & 0.727          & 0.731          & 0.623          & 0.831 \\
  thrombolysis &      & (0.802--0.822) & (0.717--0.738) & (0.720--0.741) & (0.602--0.644) & (0.818--0.845) \\
  & Fully & 0.797          & 0.724          & 0.727          & 0.633          & 0.816 \\
  &  Categorised           & (0.786--0.809) & (0.715--0.734) & (0.717--0.736) & (0.613--0.652) & (0.802--0.831) \\
\bottomrule
\end{tabular}
\end{table}

In the no recanalisation and the thrombectomy $\pm$ thrombolysis cohorts, all five metrics of the fully categorised variant overlap the 95\% CIs of the corresponding standard models. None of the paired differences reach significance, indicating that guideline-based categorisation preserves predictive performance in these two cohorts.

Categorisation produces the clearest cost in the thrombolysis cohort: AUROC, balanced accuracy, F1, and sensitivity all decrease significantly, with sensitivity showing the largest drop from 0.393 to 0.320, while specificity is unchanged (0.938 → 0.945).

Overall, the cost in predictive accuracy of guideline-based categorisation is isolated to a single cohort, leaving cohorts 1 and 3 statistically equivalent to their continuous counterparts.

\subsection{Global feature importance}
\begin{figure}[h]
  \centering
  \includegraphics[width=\linewidth]{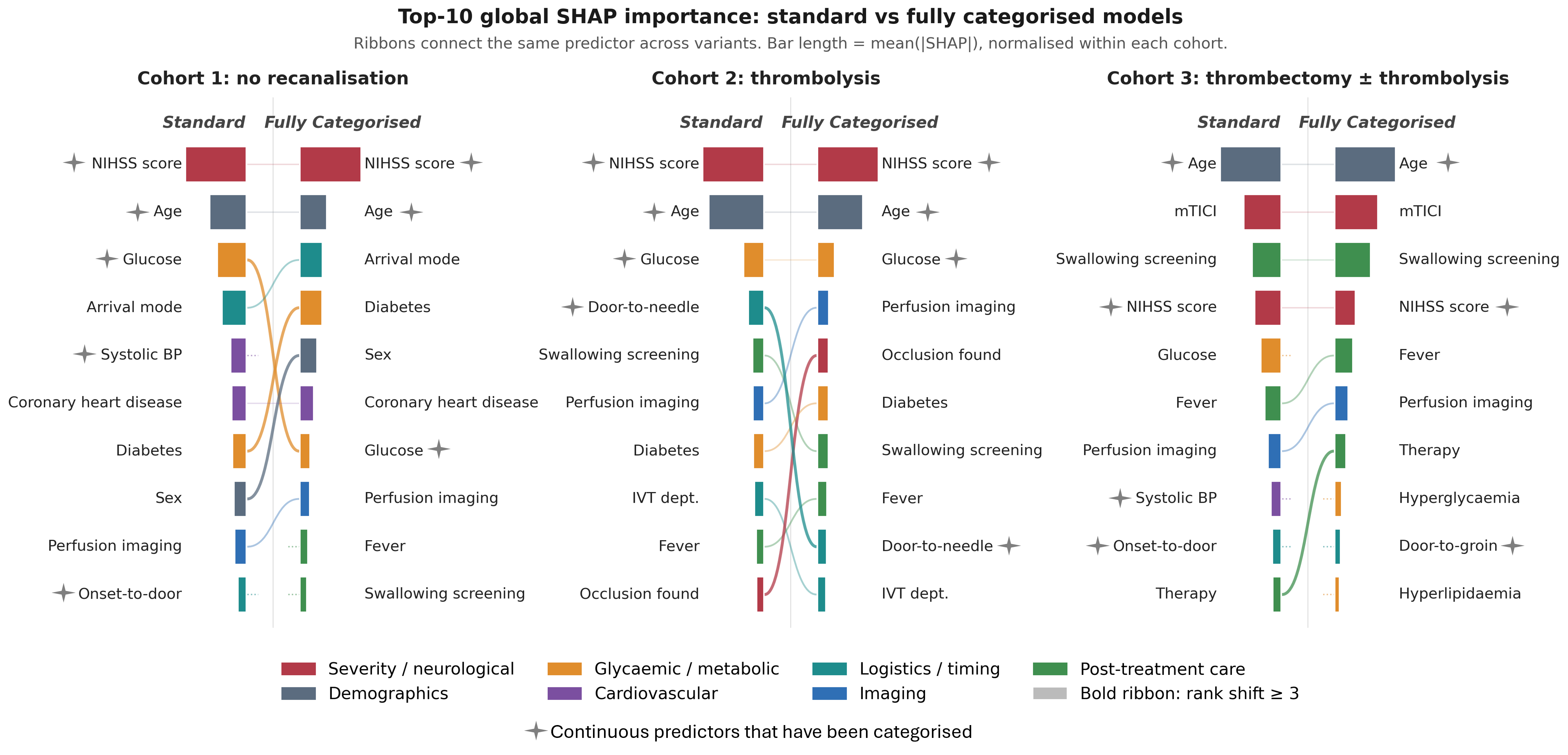}
  \caption{Top-10 global feature importances in the standard and fully categorised models for each treatment cohort.}
  \label{fig:feature_importance}
\end{figure}

For global importance ranking, we use unit-norm-normalised mean absolute SHAP values to allow direct comparison across treatment cohorts and across the standard / fully categorised variants. Figure~\ref{fig:feature_importance} shows the top-10 predictors for the standard and fully categorised models in each cohort.

The two model variants produce largely similar feature-importance ratings. NIHSS and age are among the top predictors in all six models; glucose, diabetes risk, swallowing screening, perfusion imaging, and post-treatment fever recur across cohorts. The top-10 overlap between the two model variants is 8/10 in cohort 1, 10/10 in cohort 2, and 7/10 in cohort 3, with top 2, 3, and 4 predictors overlapping in each cohort, respectively. One of the main differences between the variants is the redistribution of importance from a continuous predictor to correlated comorbidity indicators: under categorisation, glucose drops out of the top ranks while the presence of hyperlipidaemia and diabetes becomes more prominent.

\subsection{Patient-level explanations}
Patient-level SHAP visualisations were generated under the fully categorised encoding for representative cases in each cohort. Preliminary inspection indicates good agreement between the standard and fully categorised explanations on the same patients, with attributions on categorical predictors rendered against guideline-aligned reference levels rather than continuous baselines. We do not draw substantive conclusions from these examples here. The planned next step is getting clinical feedback, in which clinicians will evaluate the fully categorised model explanations against the standard model ones on alignment with their decision criteria, perceived trust, and actionability.

\section{Discussion}
Our work provides early evidence for guideline-based categorisation of continuous predictors as a viable encoding choice for gradient-boosted 90-day mRS prediction in acute ischaemic stroke. In the no-recanalisation and the thrombectomy $\pm$ thrombolysis cohorts, the fully categorised models are statistically indistinguishable from their continuous counterparts on every metric, indicating that the prognostic information carried by continuous laboratory and workflow predictors is largely retained when those predictors are re-expressed with clinically informed thresholds. Furthermore, the global SHAP rankings between the two model variants are largely concordant, with some attribution redistributed within clinical domains (top predictors shift from continuous blood glucose to diabetes and hyperglycaemia indicators and from continuous blood pressure to heart disease indicators). These suggest that categorisation re-frames explanations without changing the underlying prognostic structure.

On the other hand, we directly observe the cost of categorisation in the thrombolysis cohort, where AUROC, balanced accuracy, F1, and sensitivity all decrease significantly. The thrombolysis cohort represents the largest but also the most class-imbalanced ($\approx$1:3 favourable to unfavourable) cohort. Note that specificity is preserved while the largest decrease occurs in sensitivity, suggesting that the fully categorised encoding is more susceptible to information loss when discriminating a minority of unfavourable outcomes. Whether a sensitivity drop of this magnitude is acceptable cannot be answered from predictive metrics alone. It depends on whether the fully categorised explanations produce measurable improvements in clinician trust, comprehension, and decision quality. We plan to address this trade-off with a clinician evaluation of the fully categorised model; for now, it is the central question we leave open here.

This study has several limitations. First, the analysis is restricted to a single registry; replication on independent stroke registries is needed before claims about generalisability can be made. Second, our cohort sizes, in particular for the no recanalisation and thrombectomy $\pm$ thrombolysis cohorts (n=512 and 524), are modest, and we plan to repeat the analysis on a substantially larger sample drawn from the same registry as more data become available. A larger sample may sharpen estimates and shift the thrombolysis cohort result in either direction: the differences we currently flag as significant could attenuate as the minority class is better represented, or they could persist and become more precisely characterised. Finally, we have not yet evaluated the patient-level explanations under the new encoding with clinicians; the qualitative claim that fully categorised SHAP plots are easier to reason about than continuous ones rests, for the moment, on prior user-study evidence rather than on direct evaluation of the present models. Closing that loop with a structured comparison of standard and fully categorised individual explanations on the same patients with stroke neurologists is the planned follow-up to the present work.

\begin{acknowledgments}
This work was supported by the RESQ+ project funded by
EU’s Horizon Europe research and innovation programme under grant agreement No. 101057603.
\end{acknowledgments}

\section*{Declaration on Generative AI}
 During the preparation of this work, the author(s) used Claude Sonnet 4.6 and Grammarly in order to: Paraphrase and reword, grammar and spelling check, and generate images. After using these tool(s)/service(s), the author(s) reviewed and edited the content as needed and take(s) full responsibility for the publication’s content.
\appendix

\bibliography{bibfile}

@String{Computing = "Computing" }

@String{Springer = "Springer-Verlag" }

@inproceedings{lundberg_2017,
author = {Lundberg, Scott M. and Lee, Su-In},
title = {A unified approach to interpreting model predictions},
year = {2017},
booktitle = {Proceedings of the 31st International Conference on Neural Information Processing Systems},
pages = {4768–4777},
numpages = {10},
series = {NIPS'17}
}

@article{diprose_2024,
	title = {Abstract {TP144}: {Deep} {Learning} on {Pre}-{Procedural} {CT} and {Clinical} {Data} {Predicts} {Outcome} {Following} {Stroke} {Thrombectomy}},
	volume = {55},
	doi = {10.1161/str.55.suppl_1.TP144},
	journal = {Stroke},
	author = {Diprose, James and Diprose, William and Chien, Tuan-Yow and et al.},
	month = feb,
	year = {2024},
}

@article{abujaber_2025,
	title = {Machine learning-based prediction of 90-day prognosis and in-hospital mortality in hemorrhagic stroke patients},
	volume = {15},
	doi = {10.1038/s41598-025-90944-x},
	journal = {Scientific Reports},
	author = {Abujaber, Ahmad A. and Albalkhi, Ibrahem and Imam, Yahia and et al.},
	month = may,
	year = {2025},
	pages = {16242},
}

@article{chen_2025,
	title = {Development and validation of explainable machine learning models for predicting 3-month functional outcomes in acute ischemic stroke: a {SHAP}-based approach},
	volume = {16},
	doi = {10.3389/fneur.2025.1678815},
	journal = {Frontiers in Neurology},
	author = {Chen, Cheng-fang and Ren, Zhan-yun and Zong, Hui-hua and et al.},
	month = dec,
	year = {2025},
	pages = {1678815},
}

@inproceedings{sivaraman_2023,
	address = {Hamburg Germany},
	title = {Ignore, {Trust}, or {Negotiate}: {Understanding} {Clinician} {Acceptance} of {AI}-{Based} {Treatment} {Recommendations} in {Health} {Care}},
	doi = {10.1145/3544548.3581075},
	booktitle = {Proceedings of the 2023 {CHI} {Conference} on {Human} {Factors} in {Computing} {Systems}},
	publisher = {ACM},
	author = {Sivaraman, Venkatesh and Bukowski, Leigh A and Levin, Joel and et al.},
	month = apr,
	year = {2023},
	pages = {1--18},
}

@article{kim_2024,
	title = {Human-centered evaluation of explainable {AI} applications: a systematic review},
	volume = {7},
	doi = {10.3389/frai.2024.1456486},
	journal = {Frontiers in Artificial Intelligence},
	author = {Kim, Jenia and Maathuis, Henry and Sent, Danielle},
	month = oct,
	year = {2024},
	pages = {1456486},
}

@incollection{karagoz_2024,
	title = {Evaluating {How} {Explainable} {AI} {Is} {Perceived} in the {Medical} {Domain}: {A} {Human}-{Centered} {Quantitative} {Study} of {XAI} in {Chest} {X}-{Ray} {Diagnostics}},
	volume = {14812},
	doi = {10.1007/978-3-031-67751-9_8},
	booktitle = {Trustworthy {Artificial} {Intelligence} for {Healthcare}},
	publisher = {Springer Nature Switzerland},
	author = {Karagoz, Gizem and Van Kollenburg, Geert and Ozcelebi, Tanir and Meratnia, Nirvana},
	year = {2024},
	pages = {92--108},
}

@inproceedings{panigutti_2022,
	title = {Understanding the impact of explanations on advice-taking: a user study for {AI}-based clinical {Decision} {Support} {Systems}},
	doi = {10.1145/3491102.3502104},
	booktitle = {{CHI} {Conference} on {Human} {Factors} in {Computing} {Systems}},
	publisher = {ACM},
	author = {Panigutti, Cecilia and Beretta, Andrea and Giannotti, Fosca and Pedreschi, Dino},
	month = apr,
	year = {2022},
	pages = {1--9},
}

@article{hur_2025,
	title = {Comparison of {SHAP} and clinician friendly explanations reveals effects on clinical decision behaviour},
	volume = {8},
	doi = {10.1038/s41746-025-01958-8},
	journal = {npj Digital Medicine},
	author = {Hur, Sujeong and Lee, Yura and Park, Joongheum and et al.},
	month = sep,
	year = {2025},
	pages = {578},
}

@article{royston_2006,
	title = {Dichotomizing continuous predictors in multiple regression: a bad idea},
	volume = {25},
	doi = {10.1002/sim.2331},
	journal = {Statistics in Medicine},
	author = {Royston, Patrick and Altman, Douglas G. and Sauerbrei, Willi},
	month = jan,
	year = {2006},
	pages = {127--141},
}

@article{naggara_2011,
	title = {Analysis by {Categorizing} or {Dichotomizing} {Continuous} {Variables} {Is} {Inadvisable}: {An} {Example} from the {Natural} {History} of {Unruptured} {Aneurysms}},
	volume = {32},
	doi = {10.3174/ajnr.A2425},
	journal = {American Journal of Neuroradiology},
	author = {Naggara, O. and Raymond, J. and Guilbert, F. and et al.},
	month = mar,
	year = {2011},
	pages = {437--440},
}

@article{maslove_2013,
	title = {Discretization of continuous features in clinical datasets},
	volume = {20},
	doi = {10.1136/amiajnl-2012-000929},
	journal = {Journal of the American Medical Informatics Association: JAMIA},
	author = {Maslove, David M. and Podchiyska, Tanya and Lowe, Henry J.},
	month = may,
	year = {2013},
	pages = {544--553},
}

@article{chen_2019,
	title = {A novel approach to determine two optimal cut-points of a continuous predictor with a {U}-shaped relationship to hazard ratio in survival data: simulation and application},
	volume = {19},
	doi = {10.1186/s12874-019-0738-4},
	journal = {BMC Medical Research Methodology},
	author = {Chen, Yimin and Huang, Jialing and He, Xianying and et al.},
	month = dec,
	year = {2019},
	pages = {96},
}

@article{tustumi_2022,
	title = {Choosing the most appropriate cut-point for continuous variables},
	volume = {49},
	doi = {10.1590/0100-6991e-20223346-en},
	journal = {Revista do Colégio Brasileiro de Cirurgiões},
	author = {Tustumi, Francisco},
	year = {2022},
	pages = {e20223346},
}

@article{roe_2020,
	title = {Feature engineering with clinical expert knowledge: {A} case study assessment of machine learning model complexity and performance},
	volume = {15},
	doi = {10.1371/journal.pone.0231300},
	journal = {PLOS ONE},
	author = {Roe, Kenneth D. and Jawa, Vibhu and Zhang, Xiaohan and et al.},
	month = apr,
	year = {2020},
	pages = {e0231300},
}

@article{mortensen_2020,
	title = {Closing the {Age} {Gap} in {Acute} {Ischemic} {Stroke} {Treatment}},
	volume = {51},
	doi = {10.1161/STROKEAHA.120.030169},
	journal = {Stroke},
	author = {Mortensen, Janne Kaergaard and Andersen, Grethe},
	month = aug,
	year = {2020},
	pages = {2279--2280},
}

@article{prabhakaran_2026,
	title = {2026 {Guideline} for the {Early} {Management} of {Patients} {With} {Acute} {Ischemic} {Stroke}: {A} {Guideline} {From} the {American} {Heart} {Association}/{American} {Stroke} {Association}},
	doi = {10.1161/STR.0000000000000513},
	journal = {Stroke},
	author = {Prabhakaran, Shyam and et al.},
	month = jan,
	year = {2026},
	pages = {STR.0000000000000513},
}

@article{adams_1999,
	title = {Baseline {NIH} {Stroke} {Scale} score strongly predicts outcome after stroke: {A} report of the {Trial} of {Org} 10172 in {Acute} {Stroke} {Treatment} ({TOAST})},
	volume = {53},
	doi = {10.1212/WNL.53.1.126},
	journal = {Neurology},
	author = {Adams, H.P. and Davis, P.H. and Leira, E.C. and et al.},
	month = jul,
	year = {1999},
	pages = {126--126},
}

@misc{dorogush_catboost_2018,
	title = {{CatBoost}: gradient boosting with categorical features support},
	doi = {10.48550/ARXIV.1810.11363},
	publisher = {arXiv},
	author = {Dorogush, Anna Veronika and Ershov, Vasily and Gulin, Andrey},
	year = {2018},

}

@article{fox_monette_1992,
 author = {John Fox and Georges Monette},
 journal = {Journal of the American Statistical Association},
 doi = {10.2307/2290467},
 pages = {178--183},
 publisher = {[American Statistical Association, Taylor & Francis, Ltd.]},
 title = {Generalized Collinearity Diagnostics},
 volume = {87},
 year = {1992}
}

@article{mikulik_registry_2017,
    title = {The registry of stroke care quality ({RES}-{Q}): {The} first nation-wide data on stroke care quality},
    volume = {381},
    doi = {10.1016/j.jns.2017.08.302},
    language = {en},
    journal = {Journal of the Neurological Sciences},
    author = {Mikulik, R. and Bar, M. and Grecu, A. and et al.},
    month = oct,
    year = {2017},
    pages = {91},
}

\appendix

\section{Summary of Predictors}
\begin{table}[h!]
\centering
\caption{Summary of predictors used to model dichotomised 90-day mRS outcomes, by treatment cohort (no recanalisation, thrombolysis, thrombectomy $\pm$ thrombolysis). Continuous predictors are reported as median (IQR) and "yes/no" predictors as the percentage of \emph{yes}. C: cohort; BP: blood pressure. }
\label{tab:predictors}
\setlength{\tabcolsep}{4pt}
\renewcommand{\arraystretch}{1.05}
\footnotesize
\begin{tabular}{@{}p{3.2cm}ccc@{\hspace{5pt}}!{\vrule}@{\hspace{8pt}}p{3.5cm}ccc@{}}
\toprule
\textbf{Predictor} & \textbf{C1} & \textbf{C2} & \textbf{C3} & \textbf{Predictor} & \textbf{C1} & \textbf{C2} & \textbf{C3} \\
\midrule
\multicolumn{4}{l}{\textit{Baseline --- admission}}                        & \multicolumn{4}{l}{\textit{Baseline --- medication}}                           \\
age (years)                            & 73 (15)      & 73 (15)      & 73 (16)       & any antiplatelets         & 25.4     & 35.5     & 23.7 \\
sex (\% male)                           & 55.4         & 52.8         & 50.4          & any anticoagulants         & 19.9     & 5.7      & 19.9 \\
NIHSS score                             & 3 (4)        & 5 (4)        & 14 (9)        & \multicolumn{4}{l}{\textit{Imaging}}                                                       \\
wakeup stroke                   & 5.9          & 4.4          & 3.2           & door-to-imaging (min)                          & 15 (23)  & 9 (10)   & 10 (10) \\
arrival mode (\%)                       &              &              &               & perfusion imaging done                  & 14.0     & 17.0     & 29.7 \\
\quad EMS from home/scene               & 80.4         & 91.8         & 91.1          & old infarcts                        & 96.8     & 95.3     & 93.6 \\
\quad private transportation            & 15.2         & 6.2          & 0.6           & occlusion found                         & 13.0     & 19.1     & 99.8 \\
\quad from stroke centre       & 2.4          & 1.6          & 7.0           & \multicolumn{4}{l}{\textit{Treatment}}                                                     \\
\quad in-hospital stroke                & 2.0          & 0.4          & 1.3           & thrombolysis                      & —        & —        & 71.0 \\
hospitalised in (\%)                    &              &              &               & IVT department (\%)                 &          &          & \\
\quad ICU / stroke unit                 & 68.6         & 99.5         & 99.4          & \quad radiology                                 & —        & 50.5     & — \\
\quad standard bed                      & 25.4         & 0.2          & 0.2           & \quad emergency                                 & —        & 27.9     & — \\
\quad monitored bed                     & 6.0          & 0.3          & 0.4           & \quad stroke unit / ICU                         & —        & 21.4     & — \\
onset-to-door (hours)                    & 11.4 (19.5)   & 1.7 (1.6)     & 1.7 (1.9)     & door-to-needle (min)$^{\ast}$                  & —        & 22 (17)  & — \\
&              &              &               & door-to-groin (min)                             & —        & —        & 67 (35) \\
\multicolumn{4}{l}{\textit{Baseline --- comorbidities}}                                    & door-to-reperfusion (min)                      & —        & —        & 102 (50) \\
atrial fibrillation             & 20.9         & 11.0         & 27.1          & mTICI (\% $>$2A)                                & —        & —        & 84.5 \\
hypertension                   & 78.9         & 74.5         & 71.5          & \multicolumn{4}{l}{\textit{Post-treatment}}                                                \\
diabetes                        & 57.4         & 47.0         & 43.8          & stroke etiology known                   & 69.6     & 72.8     & 58.5 \\
hyperlipidaemia                 & 42.8         & 42.0         & 38.1          & fever diagnosed                       & 7.3      & 6.3      & 11.2 \\
congestive heart failure        & 7.0          & 5.0          & 8.8           & hyperglycaemia diagnosed                & 14.8     & 13.4     & 13.1 \\
smoker                          & 25.8         & 27.1         & 25.6          & swallowing screening (\%)                  &          &          & \\
previous stroke                 & 16.8         & 20.4         & 13.1          & \quad yes                                       & 86.2     & 92.7     & 88.4 \\
coronary heart disease      & 14.8         & 11.9         & 14.5          & \quad no                                        & 9.5      & 3.2      & 0.4 \\
&              &              &               & \quad not applicable                            & 4.3      & 4.1      & 11.2 \\
\multicolumn{4}{l}{\textit{Baseline --- labs }}                      & therapy (\%)                              &          &          & \\
systolic BP (mmHg)               & 153 (34) & 160 (33) & 152.0 (30)  & \quad yes                                       & 81.5     & 83.7     & 84.1 \\
diastolic BP (mmHg)              & 87 (15)  & 87 (16)  & 85.0 (16)   & \quad no                                        & 6.5      & 3.7      & 8.9 \\
blood glucose (mg/dL)                      & 113 (45)   & 121 (47)   & 121 (40)    & \quad not required                              & 12.0     & 12.6     & 7.0 \\
cholesterol (mg/dL)                   & 47 (29)  & 47 (27)  & 45 (25)   &                                                  &          &          &      \\
\bottomrule
\end{tabular}
\flushleft\footnotesize $^{\ast}$~Door-to-needle time is included in door-to-groin time when thrombolysis precedes thrombectomy.
\end{table}

\end{document}